# Using Vision Language Foundation Models to Generate Plant Simulation Configurations via In-context Learning

**Heesup Yun[a], Isaac Kazuo Uyehara[b], Earl Ranario[a], Lars Lundqvist[c], Christine H. Diepenbrock[b], Brian N. Bailey[b], J. Mason Earles[a,d]***

[a] Department of Biological and Agricultural Engineering, University of California, Davis, CA 95616, USA.
[b] Department of Plant Sciences, University of California, Davis, CA 95616, USA.
[c] Department of Electrical and Computer Engineering, University of California, Davis, One Shields Avenue, Davis, CA 95616, USA.
[d] Department of Viticulture and Enology, University of California, Davis, One Shields Avenue, Davis, CA 95616, USA.
* Corresponding author: J. Mason Earles (jmearles@ucdavis.edu)

## Abstract

This paper introduces a benchmark for evaluating whether vision-language models (VLMs) can generate plant simulation configurations from imagery using in-context learning. We study this benchmark for cowpea plot reconstruction for plant simulations, where the VLM needs to generate structured JSON configurations that include field and plant information. Open-source multimodal models from Gemma 4 and Qwen3.5 families are evaluated on a synthetic cowpea dataset with known JSON ground truth and on a real drone orthophoto dataset with field-collected JSON. Five in-context learning methods are used, from format restriction instruction to few-shot image examples with auxiliary grounding information. The results show that VLMs can generate valid JSON outputs, can generally estimate days after planting (DAP), plant counts, plant locations, sun angles, and leaf chlorophyll content, and can render approximate simulations of cowpea plots. Error metrics fluctuate across model families and often remain worse than dataset baselines, particularly when VLMs' pretrained knowledge dominates over weak visual evidence. These results position image-to-simulation JSON generation as a promising but currently challenging task, and establish a benchmark for studying how multimodal reasoning, prompt design, and the sim-to-real domain gap affect plant phenotyping tasks.

## 1. Introduction

Plant modeling and simulation help analyze plant performance and interactions with the environment. They also help understand the relationship between plant genotypes and their shape, and design ideal plant types. Detailed 3D representations of the field, plant positions, and canopy architecture are essential for simulating biophysical processes such as photosynthesis and water use (Gaillard et al., 2020; Soualiou et al., 2021).

Functional-structural plant modeling (FSPM) provides a framework to simulate plant behavior based on plant architecture, considering biophysical and environmental factors, from molecular and cellular levels, to the organ level, and up to the ecological level (Soualiou et al., 2021). For FSPMs to be useful and reproducible, they require structured representations of plot geometry, environmental conditions, and crop 3D architecture, because simulation quality depends on how accurately those inputs describe the real world.

To address this need, FSPMs often use structured languages to define plants and their parameters, such as L-systems (Lindenmayer, 1968a, 1968b), XL (Kurth and Ong, 2017), and L-Py (Boudon et al., 2012). Most previous studies used a forward workflow where experts define rules and structures, from which simulations or images are then generated (Baker et al., 2024). However, to faithfully represent real-world plants, an inverse approach is needed: generating plant models from sensing data. Specifically, research on generating FSPMs from images for multiple plants at the plot or field scale is still lacking.

To bridge the gap between field-scale sensing data and plant modeling, recent studies have focused on extracting architectural measurements from sensor data. For example, Zhu et al. (2023) reconstructed organ-scale phenotypic parameters including leaf length, width, and area from 3D point clouds of field crops using unmanned aerial vehicle (UAV) imagery. Similarly, Chebrolu et al. (2021) developed methods for registering spatio-temporal point clouds to track plant traits at the organ level over time, while Ziamtsov and Navlakha (2019) used deep learning to classify plant organs and skeletonize stems from 3D point clouds with high accuracy. However, these extracted measurements still need to be transferred into structured FSPM definitions to enable functional simulations.

Recent large language models (LLMs) and vision-language models (VLMs) have shown that structured document outputs such as JSON, HTML, and tabular representations can be generated from textual or visual context (Lee et al., 2022; Zhang et al., 2023). These advancements have raised the possibility that a multimodal model could directly infer plant simulation configuration documents from plot images, rather than relying on manual reconstruction pipelines. However, the feasibility of this idea depends not only on generating valid structured outputs from image context, but also on recovering biologically and agronomically meaningful parameters from the image itself.

Agricultural VLM benchmarks have expanded rapidly to evaluate the feasibility of VLMs for agricultural perception tasks (Arshad et al., 2025; Awais et al., 2025; Shinoda et al., 2025; Yang et al., 2025; Zhou et al., 2024). Existing agricultural VLM benchmarks mainly focus on classification, identification, question answering, and related agronomic perception tasks. While these benchmarks have broadened the evaluation of VLMs in agriculture, there is still no benchmark that tests whether VLMs can generate simulation-ready plant configuration files that combine multiple tasks, such as plant localization and the estimation of environmental and biophysical traits within a structured document format. It remains unclear whether current VLMs can support image-to-simulation workflows.

To address this gap, we make three main contributions:

- A benchmark for image-to-plant-simulation JSON generation, comprising synthetic cowpea plots with known ground truth and a real drone orthophoto dataset with field-collected ground truth.[1]
- A systematic evaluation of six open-source VLMs under five in-context learning methods, quantifying JSON integrity, geometric accuracy, and biophysical parameter estimation.[2]
- An analysis of how fine-tuning, few-shot data modality, and contextual bias affect VLM performance on plant simulation generation.

The remainder of this paper is organized as follows. Section 2 reviews related work on structured document generation and agricultural VLM benchmarks. Section 3 describes the cowpea dataset, simulation pipeline, JSON schema, and the VLM evaluation protocol. Section 4 presents the results across model families and in-context learning methods. Section 5 discusses the implications of the results. Section 6 concludes with limitations and future directions.

---

[1] Dataset is publicly available here: https://huggingface.co/datasets/heesup/vlm-plant-sim
[2] Code is publicly available here: https://github.com/heesup/vlm-plant-sim

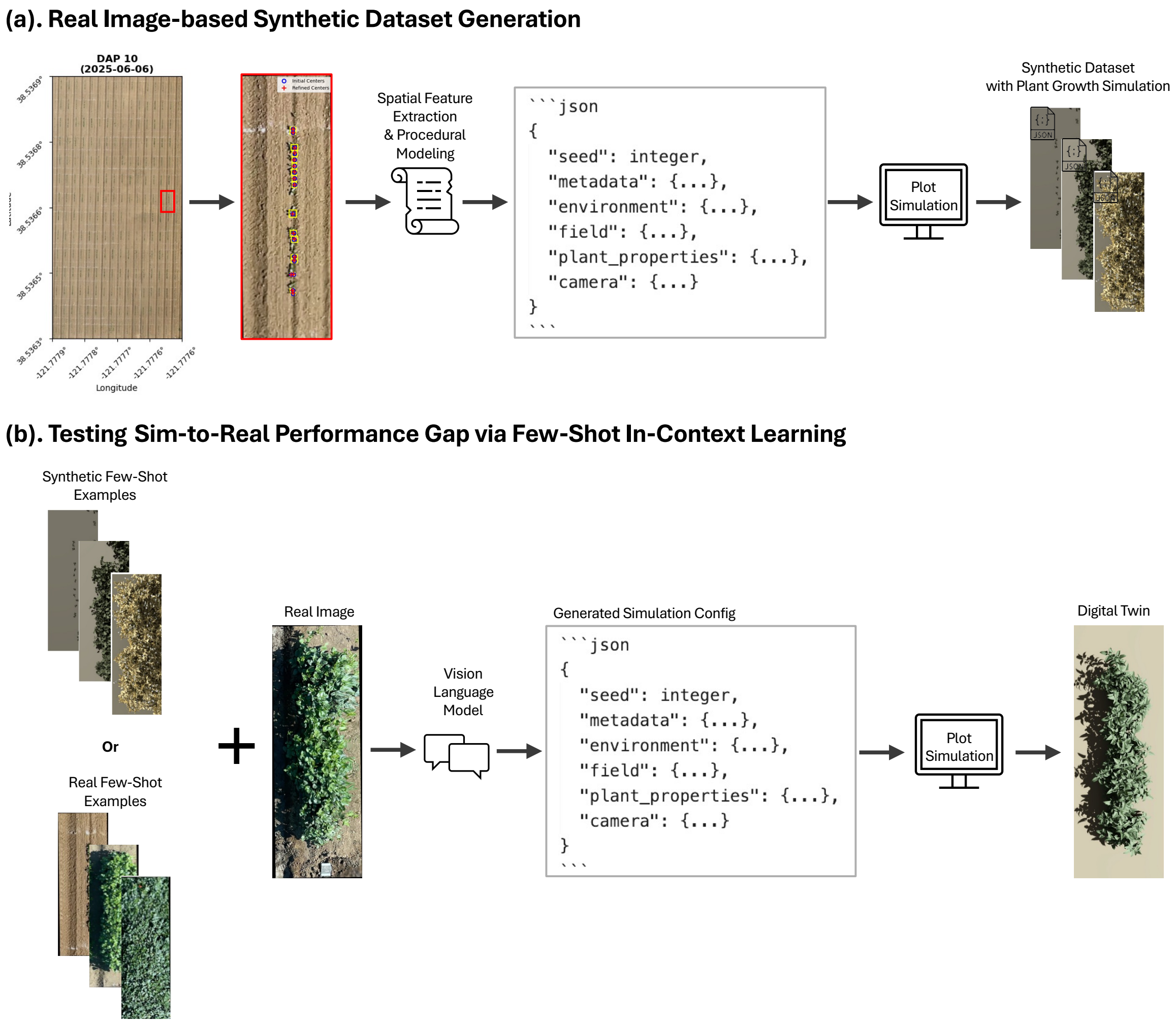

Figure 1. Overall diagram of generating datasets and evaluating vision-language models for plant simulation generation. (a) Synthetic Dataset Construction: Real drone orthophotos undergo plant detection and spatial feature extraction to generate structured JSON configurations, which drive procedural plot simulations across growth stages (10–90 DAP). (b) Sim-to-Real Evaluation: Vision-language models receive few-shot examples (synthetic or real) paired with a target real image to generate simulation configurations via in-context learning, enabling cross-domain performance assessment.

## 2. Related Work

### 2.1. Structured Document Formats for Plant Modeling

Structured languages have been actively used in plant modeling to encode architectural rules and plant developmental processes. L-systems provide the foundational grammar for plant development through parallel string rules (Lindenmayer, 1968a, 1968b), implemented in L-Studio (Karwowski and Prusinkiewicz, 2004). L-Py (Boudon et al., 2012) and XL (Kurth and Ong, 2017) extend L-systems to have modular model development, cross-scale and hierarchical plant model representations. In addition to modeling the plants, structured document formats facilitate diverse plant simulation tasks. For example, Lobet et al. (2015) used XML to represent root architectural data, and Baker et al. (2024) used JSON to interface between FSPMs and Unreal Engine for visualizations.

## 2.2. Structured Document Generation with LLMs and VLMs

Recent advances in VLMs have enabled models to understand the schema and content of structured documents and to generate them from a given context (Dong et al., 2024; Geng et al., 2023; Yang et al., 2026; Zhang et al., 2023). Furthermore, VLMs can analyze images and generate responses, not only explaining their contents but also counting and localizing objects within them (Kang et al., 2024; Paiss et al., 2023; Radford et al., 2021; Subramanian et al., 2022; Xu et al., 2016). By combining this vision capability with LLMs' structured document generation, VLMs can generate structured outputs from an image. Kim et al. (2022) proposed a method that converts an image into a JSON text document using an image encoder and text decoder, without optical character recognition. Lee et al. (2022) proposed Pix2Struct, which converts a webpage screenshot into a structured HTML document. Liu et al. (2023a,b) suggested DePlot and MatCha for translating chart or graph images into a structured Markdown table or a structured dataset.

## 2.3. Agricultural VLM Benchmarks and Limitations for Plant Simulations

Driven by the zero-shot image-to-text capabilities of state-of-the-art VLMs, specialized agricultural benchmarks have emerged to evaluate tasks such as disease identification and quantification. Zhou et al. (2024) proposed a vision-language model for crop disease classification (VLCD) for few-shot crop leaf disease classification using Qwen-VL-generated prompts. Yang et al. (2025) trained AgriGPT-VL on Agri-3M-VL and evaluated it on AgriBench-VL-4K, which contains agricultural open-ended visual question answering (VQA) and image-grounded multiple choice question answering (MCQA). Awais et al. (2025) trained AgroGPT on the AgroInstruct dataset to enable multi-turn multimodal dialogue about agricultural concepts (e.g., diseases, weeds, insects, fruits), and assessed VQA using the AgroEvals benchmark. Shinoda et al. (2025) introduced AgroBench (Agronomist AI Benchmark), an expert-annotated multiple-choice question answering (MCQA) benchmark that evaluates VLMs' performance on seven agricultural tasks: three identification tasks (disease, pest, and weed), three management tasks (crop, disease, and traditional), and a machine usage question answering task. Arshad et al. (2025) introduced the AgEval benchmark for the identification, classification, and quantification of plant disease and stress levels.

Existing agricultural VLM benchmarks evaluate independent perception tasks, such as disease classification, visual question answering, or object detection. Our benchmark requires VLMs to simultaneously perform plant localization, lighting estimation, biophysical parameter estimation, and structured document generation within a single JSON output. This multi-task integration introduces two challenges absent in prior work: (1) cross-parameter dependencies, such as errors in lighting estimation propagating to chlorophyll predictions, and (2) structured output constraints, where syntactically valid JSON must also contain geometrically and biologically meaningful values. Consequently, our work differs from prior agricultural VLM benchmarks not merely in task novelty, but in testing whether vision-language models can serve as end-to-end simulation configuration generators rather than perception-only modules.

# 3. Methods

## 3.1. Drone Remote Sensing Dataset

Drone-based remote sensing of a cowpea (Vigna unguiculata (L.) Walp.) field was conducted in 2025 at a breeding trial field in California. The plot dimensions were 1.5 m by 3.0 m with 1.5 m alleys separating each plot, and the plants were planted in a single row within the plot. The field had 40 beds and 14 tiers with borders to minimize edge effects. 60 cowpea genotypes were planted from beds 2 to 16 with three replicate blocks, and the rest of the field was planted with common bean and lima bean populations. The cowpeas were planted on May 27, 2025, and harvested on October 9, 2025.

Drone imagery was collected in 2025 using a custom ArduPilot-based platform equipped with an iPhone 13 Pro mounted on 3D-printed attachments, flown at 10 m above ground level. The 20-minute flight missions were conducted on 10, 28, 49, 70, and 87 days after planting (DAP), with 80% side overlap and 70% frontal overlap. The orthophotos were processed with OpenDroneMap and exported as GeoTIFF files. Plot boundaries were annotated using QGIS, and the plot images were cropped accordingly.

Ground truth DAP values were calculated as the difference between the drone data collection date and the planting date. Plant counts and locations were manually annotated at 10 DAP images using LabelMe software, saved in common objects in context (COCO) JSON format. These annotations were converted to plant center coordinates (in meters), and used as ground truth for the rest of the season. Because individual plants remained rooted in fixed positions after emergence, as shown in Figure 2, we used these center locations as reference geometry across later dates.

The sun elevation and azimuth angles at the time of image capture were calculated in degrees using the pvlib Python library (Anderson et al., 2023), with the plot center's latitude and longitude and the exact image capture timestamp. Ground truth chlorophyll values were sampled three times per season in 24 of the 180 cowpea plots using a chlorophyll content index (CCI) meter (Apogee MC100, Apogee Instruments, Inc., USA) and converted from CCI units to chlorophyll content (μg/cm²) using an empirical equation derived from a previous study (Parry et al., 2014). The measured time series of chlorophyll content per plot was fitted to polynomial functions of DAP and resampled for the drone data collection dates. The DAP, plant count, plant locations, sun elevation, sun azimuth, and chlorophyll were used as the benchmark dataset for further evaluation.

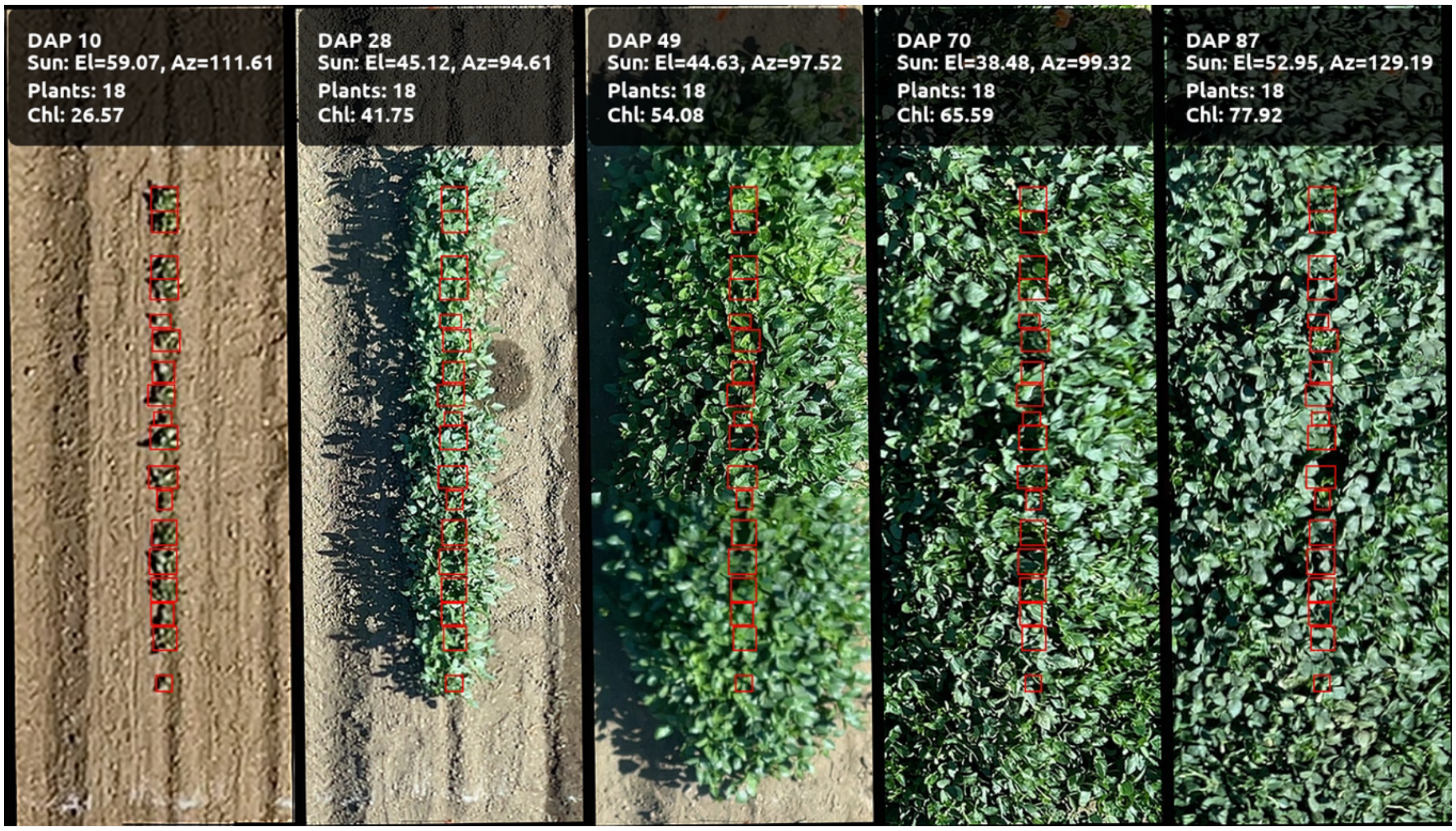

Figure 2. Temporal progression of a representative cowpea plot (bed 2, tier 2) across the growing season. Images show canopy development from early emergence (10 DAP) to maturity (87 DAP), with red bounding boxes indicating initial plant locations tracked throughout the season. Days after planting (DAP), sun elevation and azimuth angles (El/Az) were calculated from image metadata. Plant count and locations were annotated from the DAP 10 image. Leaf chlorophyll content (Chl) was derived from polynomial fittings of the field measurement.

### 3.2. Cowpea Plot Simulation JSON

A structured JSON schema was used to capture the minimum information required to reconstruct a cowpea plot in the simulation environment. The schema includes metadata, environment, field layout, plant properties, and camera parameters. Some fields are visually inferable from imagery, such as DAP, whereas others are metadata, such as year and location. The JSON file has six top-level fields: random seed, metadata, environment, field, plant properties, and camera setup.

The JSON file starts with a random seed for reproducibility, which controls the random sampling used to procedurally generate plants and sample the parameters in the simulation. The metadata key includes year, location, plant type, and days after planting. The environment key contains soil and sun environment data, such as soil spectral data categories and specular reflection coefficients, as well as sun elevation and azimuth degrees. The field key specifies the field layout, including plot size and the number of beds. Plot keys are nested under the field key, and the plot key contains plot bed and row ID, and individual plant locations as an array. The plant properties key includes a chlorophyll parameter that determines the leaf color spectrum. Image sensor and camera positioning related parameters are in the camera key, such as shutter speed, ISO, camera resolution and model, camera height, and a look-at vector.

### 3.3. Cowpea Plot Simulator

A plant simulator program was developed using the Helios 3D simulation library (Bailey, 2019) to generate synthetic cowpea plot images from plant simulation JSON files. The program generates a synthetic cowpea plot with full 3D plant geometry by reading initial plant locations, DAP, solar position, chlorophyll content, and camera parameters. The Helios library initializes plants on the specified plant locations, and advances time forward to grow the plant architecture procedurally up to the desired DAP based on phenological stages. The plant architectural growth is simulated with phyllochron control, which is the minimum number of days between successive phytomer emergence along a shoot (Bailey, 2025). All the growth and shape parameters were inherited from default plant architecture library settings.

To colorize the generated 3D plant architecture model, leaf and stem spectra were generated from the PROSPECT leaf optics model (Jacquemoud and Baret, 1990), while flower and pod spectra were generated from a mixture of standard purple and yellow spectra. To simulate various lighting conditions, sun elevation and azimuth were randomly sampled from uniform distributions over [0°, 90°] and [0°, 360°], respectively.

## 3.4. Synthetic Cowpea Dataset

To generate synthetic data following real-world field configurations in a scalable way, an automated pipeline was developed. Figure 1 (a) shows the overview of the synthetic image generation process. The Excess Green (ExG) (Woebbecke et al., 1995) vegetation index was applied to segment plants from the background, and plant blobs were detected using OpenCV blob detection functions. The detected blobs were divided into individual plants based on the blobs' aspect ratio, and the center positions were refined to the center of gravity of each blob. The plant locations were calculated considering the pixel locations and the plot dimensions in meters, with the image center set to (0,0).

The plant stand count and plant locations were detected from the drone orthophoto from the 10 DAP across 560 plots, and the detected plant locations were saved in JSON format to simulate the cowpea plants in the plant simulator. Starting from 0 DAP, cowpea plant growth simulations were performed for 10, 30, 50, 70, and 90 DAP, resulting in a total of 2,800 synthetic cowpea plot simulations. The simulated plots were exported to JSON ground truth files and corresponding JPEG images in 381 × 1080 resolution.

### 3.5. Vision-Language Models

To evaluate the VLMs' ability to generate a simulation config JSON file from an image, two families of vision-language models were tested. We selected Gemma 4 (Google DeepMind, 2026) and Qwen3.5 (QwenTeam, 2026) because they represent recent open-source multimodal model families with various model sizes, strong instruction-following and reasoning abilities, and practical support for controlled local inference without API costs. Gemma 4 was released in April 2026 and is designed to run on edge devices, with model sizes ranging from 2B to 31B parameters. Qwen3.5 was released in February 2026 and includes models from 0.8B to 122B parameters. We tested the 2B, 4B, and 31B Gemma 4 models and the 4B, 9B, and 27B Qwen3.5 models. A self-hosted vLLM (Kwon et al., 2023) server was used to provide an API endpoint for interacting with the models via a Python script. The maximum context size was set to 32K to accommodate the maximum length of in-context learning methods and generation results, and the temperature was set to 0 for deterministic generation.

### 3.6. In-context Learning

LLMs can learn from context and provide an answer based on the given information without updating the model weights, a capability called few-shot learning or in-context learning (Brown et al., 2020). We tested in-context learning methods by progressively providing context to the model and observing changes in JSON generation performance. A summary and examples of the five in-context learning methods are shown in Table 1.

The simplest approach is to provide format restriction instructions (FRI), such as “Respond as JSON" or “Provide your output in the following valid JSON format”. Therefore, the first method (M1) served as a baseline that defined the LLMs' role, task, goal, and JSON FRI. The second method (M2) added a JSON schema to the first, so the models could reference the JSON structure and variable types. The third method (M3) added a few-shot JSON examples to the second method. Three JSON files were selected for 10, 50, and 90 DAP for the synthetic dataset, and 10, 49, and 87 DAP from the real dataset. For each selected DAP, samples that maximized the variance in plant count, sun elevation, sun azimuth, and chlorophyll were selected.

An important element from the example JSON was the reasoning key text, which provided references for building step-by-step reasoning text to encourage intermediate decomposition of the tasks. Following this, VLMs were instructed to include the ‘reasoning’ field before producing the structured output. This two-stage process enables the model to reason freely in natural language, then complete the remaining JSON structure based on the analysis (Tam et al., 2024). The reasoning field was treated as a prompting aid to help the model rather than as a validated explanation of the models' reasoning and was not included in the evaluation metrics.

The fourth method (M4) included few-shot image-JSON pairs, so that the models could learn how to generate JSON from images by extracting visual context. Three pairs of images and JSON were delivered as a list of messages between the user and the assistant, simulating three pairs of questions and answers.

The fifth method (M5) provided the model with auxiliary grounding hints. M5 added exact DAP, solar positions, and relative plant locations to M4, but absolute plant locations and chlorophyll content were not included. We designed M5 to read DAP, sun elevation, and azimuth from the grounding text, convert relative plant locations to absolute locations, and estimate chlorophyll content from the image without any hint. This method tests whether VLMs can reliably integrate structured hints in plain text into a valid JSON simulation configuration and it assesses the language performance of the models and how chlorophyll content estimation accuracy changes if the rest of the parameters were given.

Table 1. In-context learning configurations and example texts added to the prompt.

| Configuration | Example |
|---|---|
| M 1:<br><br>Zero-shot JSON generation | You are a plant phenotyping expert analyzing Helios 3D plant simulator images. Extract all simulation parameters that produced this image and output them as JSON.<br><br>--- Parameter Reference ---<br>1. metadata<br><br>…(more)<br><br>Answer: |
| M 2:<br><br><br>M 1<br>+ JSON schema | JSON SCHEMA:<br>```json<br>{<br>  "reasoning": "string",<br>  "seed": "integer",<br>  "metadata": {<br>    "year": "integer",<br>    "location": "string",<br>    "plant_type": "string",<br>    "dap": "integer"<br>  },<br>   …(more)<br><br>  } |
| M 3:<br><br><br>M 2<br>+ few-shot JSONs | Example 1:<br>```json<br>{<br>  "reasoning": "Visual analysis: Plant maturity suggests 10 days growth. …(more)   x 3 |
| M 4:<br>M 3<br>+ few-shot images | Example 1:<br>user: <IMAGE><br>assistant:<br>```json<br>{   x 3<br>  "reasoning": "Visual analysis: Plant maturity suggests 10<br><br>days growth. …(more) |
| M 5:<br><br><br>M 4<br>+ grounding info | Ground truth hints for target image:<br>- Plant age: 10 DAP<br>- Plant count: 14 plants<br>- Sun position: 62.9 deg elevation, 169.4 deg azimuth<br>- Plant approximate relative locations [0.0-1.0] (rx, ry): [(0.163, 0.073), (0.162, 0.085), (0.56, 0.367), (0.556, 0.38), (0.552, 0.558), (0.563, 0.579), (0.564, 0.605), (0.568, 0.648), (0.561, 0.657), (0.567, 0.699), (0.568, 0.722), (0.563, 0.745), (0.575, 0.768), (0.174, 0.743)]<br>  Convert to meters: x = (rx - 0.5) * 1.3521, y = -(ry - 0.5) * 3.8405 |

### 3.7. Evaluation Metrics

The JSON output quality was evaluated in three categories: JSON integrity, geometric accuracy, and lighting/biophysical accuracy. All six models were evaluated on a synthetic cowpea dataset, and the few-shot examples provided to the models were excluded from the evaluation.

**JSON integrity:** We evaluated JSON integrity metrics to assess quality differences in JSON generation across models and contexts, focusing on structural reliability of generated JSON files. The first metric was the JSON syntax error rate, the percentage of the response containing JSON syntax errors that could not be parsed into JSON,

$$\text{Syntax Error Rate (\%)} = \left(1 - \frac{1}{N}\sum_{i=1}^{N} \mathbb{1}_{\text{valid}}(x_i)\right) \times 100\,,$$

here $N$ is the total number of generated responses, $x_i$ is the $i$-th generated response, and $\mathbb{1}_{\text{valid}}(x_i)$ is an indicator function that equals 1 if $x_i$ can be successfully parsed into a valid JSON dictionary, and 0 otherwise.

The second metric was the JSON key-missing rate, which counts the number of missing keys in the response and divides by the total number of JSON keys in the ground truth,

$$\text{Key Missing Rate (\%)} = \frac{1}{|V|}\sum_{v \in V} \left(1 - \frac{K_{\text{present}}^{(v)}}{K_{\text{expected}}^{(v)}}\right) \times 100,$$

where $V$ is the set of all successfully parsed JSON responses (where $\mathbb{1}_{\text{valid}}(x_i) = 1$), $K_{\text{expected}}^{(v)}$ is the total number of expected keys (including nested properties) defined in the ground-truth schema for response $v$, and $K_{\text{present}}^{(v)}$ is the number of those expected keys that are successfully found in the generated response $v$.

**Geometric evaluations:** The geometric evaluation metrics represent plant geometric errors indirectly, by comparing the DAP, plant count, and plant locations. Because DAP caused major geometric changes in a cowpea plot by altering plant size, architecture, and plant organ visibility, as shown in Figure 2, DAP was categorized in the geometric evaluations category. The DAP and plant count errors were evaluated using mean absolute error (MAE).

$$\mathrm{MAE} = \frac{1}{N}\sum_{i=1}^{N}|y_i - \hat{y_i}|$$

For plant locations, spatial alignment between the predicted $S_1$ and ground truth $S_2$ plant locations was calculated using Chamfer distance (CD) (Barrow et al., 1977),

$$\mathrm{d_{CD}} = \mathrm{d}(\mathrm{S_1}, \mathrm{S_2}) + \mathrm{d}(\mathrm{S_2}, \mathrm{S_1}),$$

where $\mathrm{d}(P,\ Q)$ denotes the average nearest-neighbor distance from set $P$ to $Q$.

**Lighting and biophysical evaluations:** Sun elevation, azimuth, and chlorophyll content change the lighting and color of the cowpea plots. Sun elevation and azimuth affect shadow length, shadow direction, and canopy illumination, whereas chlorophyll content changes leaf color and apparent greenness. For sun elevation, sun azimuth, and chlorophyll content, MAE was calculated.

## 3.8. Statistical Analysis and Dataset Baselines

The effects of model family, model size, and in-context learning method were analyzed statistically. Bar graphs with 95% confidence intervals (CIs) and statistical groups were plotted for JSON syntax error rate, key-missing rate, and parameter MAEs. The CIs for JSON syntax error and key-missing rate were computed using a Bernoulli distribution, while CIs for continuous metrics were computed using a normal distribution.

Statistical differences between models within the same in-context learning setting were assessed using the Kruskal–Wallis H-test, followed by pairwise Mann–Whitney U tests with Bonferroni correction. Statistical groups were displayed when there were significant differences between the groups. Additionally, for the largest model from Gemma 4 and Qwen3.5 model families, separate bar graphs were plotted grouped by DAP, with CIs and statistical groups indicated. Median guess baseline ($B_{med}$) and random guess baseline ($B_{rnd}$) were calculated as dataset baselines. $B_{med}$ represents performance when estimating values using only marginal distributions of parameters without image understanding. It was obtained by predicting each parameter using the median of its valid range and computing MAE against the ground truth. $B_{rnd}$ represents randomized guessing. To reduce stochastic variability, $B_{rnd}$ was estimated via Monte Carlo simulation by sampling 1,000 values from each parameter's distribution and calculating the MAE relative to ground truth.

For plant locations, modified versions of $B_{med}$ and $B_{rnd}$ were used because MAE is inappropriate for spatial coordinates. First, the median plant count was determined from the ground truth distribution and used as the number of clusters. All plant locations were pooled and K-means clustering was used to partition them into clusters. $B_{med}$ was then calculated as the CD between the resulting cluster centers and dataset plant locations, averaged across samples. For $B_{rnd}$, plant x and y coordinates were randomly sampled from plant location ranges, CDs were calculated against ground truth, and the results were averaged.

## 3.9. Vision Foundation Model Baselines

To compare few-shot learning capability between VLM and vision foundation model (VFM), VFM baselines were calculated using linear models fitted to VFM embeddings. For the VFM baseline, a linear model using zero-shot VFM embeddings was fitted to three few-shot images, yielding a minimum-norm solution. A VFM, DINOv3 (Siméoni et al., 2025) trained on a satellite dataset was used, providing 1024 embedding dimensions. Let $X \in \mathbb{R}^{N \times D}$ denote the embedding matrix extracted from the VFM, where N=3 and D=1024. An augmented feature matrix $\tilde{X} = [X, 1_N] \in \mathbb{R}^{N \times D+1}$ was defined, where $1_N$ is a column vector of ones. The linear mapping to the target plant simulation parameters $Y \in \mathbb{R}^{N \times K}$ is given by $Y = \tilde{X}W$. $W \in \mathbb{R}^{(D+1) \times K}$ was the augmented weight matrix combining both the projection weight and the bias, and K=5 for DAP, plant count, sun elevation, sun azimuth, and chlorophyll content.

Given that the number of few-shot samples N=3 was smaller than the feature dimension D=1024 ($N \ll D + 1$), the linear system was highly underdetermined, yielding an infinite number of exact solutions that satisfy zero fitting error. To prevent overfitting and ensure a unique and stable solution, we constrained the hypothesis space, seeking the minimum-norm solution by $\min_{W} ||W||_F^2$ subject to $Y = \tilde{X}W$ where $||\cdot||_F$ denotes the Frobenius norm. The optimal closed-form solution to this constrained optimization was obtained by using the Moore-Penrose pseudo-inverse $\tilde{X}^+$ (Golan, 2012), $W = \tilde{X}^+ Y$. Since the few-shot samples were sampled independently, rows $\tilde{X}$ were linearly independent, so the pseudo-inverse can be explicitly computed as $\tilde{X}^+ = \tilde{X}^T \left(\tilde{X}\tilde{X}^T\right)^{-1}$. Therefore, the weights matrix was computed by $W = \tilde{X}^T \left(\tilde{X}\tilde{X}^T\right)^{-1} Y$.

## 3.10. Effect of Dataset Modalities on Fine-Tuning and Few-Shot In-Context Learning

To evaluate the effect of fine-tuning and few-shot modalities on parameter estimation, we calculated Pearson's correlation coefficient ($r$), MAE, and symmetric mean absolute percentage error (sMAPE),

$$\text{sMAPE (\%)} = \frac{1}{N}\sum_{i=1}^{N}\frac{200 \cdot |y_i - \widehat{y_i}|}{|y_i| + |\widehat{y_i}|}$$

where $N$ is the total number of valid samples, $y_i$ is the ground truth parameter value for the $i$-th sample, and $\widehat{y_i}$ is the predicted parameter value for the $i$-th sample.

**Effect of Synthetic Fine-Tuning on Synthetic and Real Datasets Evaluation:** To test whether fine-tuning on synthetic data could improve structured generation beyond in-context learning, the largest model from each family was fine-tuned on synthetic image-JSON pairs. Therefore, we focused on the largest models because they showed the strongest in-context learning performance and were the most plausible candidates for practical application. To fine-tune the model on a small GPU cluster, parameter-efficient fine-tuning (PEFT) was performed using low-rank adaptation (LoRA) (Hu et al., 2021) with r=16 and α=16. The Qwen3.5 27B and Gemma 4 31B models were fine-tuned with approximately 124M and 133M trainable parameters, respectively. Data from beds 9 to 40 of the synthetic cowpea dataset were split 80/20 for training and validation, excluding beds 1 to 8, which were reserved for the benchmark dataset. We used the Unsloth (Han et al., 2023) Python library to accelerate training and reduce memory overhead. Each model was trained for 5 epochs with an effective batch size of 32 across 4 NVIDIA H100 GPUs for about 10 hours per model.

**Effect of Synthetic and Real Few-Shot Contexts on Real Dataset Evaluation:** Another scenario involved evaluating the base models with different few-shot example modalities. The key question was whether real few-shot context improves inference on the real dataset more than synthetic few-shot context. To test this, the largest model per model family was evaluated on the real dataset under two few-shot context modalities, synthetic few-shot examples and real few-shot examples, as shown in Figure 1 (b).

## 4. Results

### 4.1. Synthetic Dataset Evaluation Result

Figure 3 shows the evaluation results across two model families (Gemma 4 and Qwen3.5), three model sizes (2B, 4B, and 31B for Gemma 4; 4B, 9B, and 27B for Qwen3.5), and five in-context learning methods (M1–M5). Across the synthetic benchmark, the models were generally capable of producing near-complete JSON outputs, especially after a schema was provided. However, this structural reliability did not ensure accurate parameter estimation. Performance depended strongly on model family, model size, and in-context learning methods. Several combinations of these factors remained worse than simple prior-based baselines for plant count, plant location, and biophysical variables.

**JSON Integrity Evaluations:** The Gemma 4 2B model showed the highest 2.68% JSON syntax error rate for M2, while the Qwen3.5 4B model showed the maximum 3.76% for M3. Adding few-shot JSON examples increased the JSON syntax error rate for the Qwen3.5 4B model. The Gemma 4 2B model with M1 showed 1.13% key-missing rate, while the Qwen3.5 9B showed 0.12% on the same method. For all models, adding a JSON schema (M2–M4) reduced the key-missing rate to 0.0%.

**Geometric Evaluations:** The Gemma 4 models generally showed decreasing DAP MAEs as model size increased. The Gemma 4 models showed lower DAP MAE values for M1, M2, and M4 than Qwen3.5 for all model sizes. The Qwen3.5 9B model showed the highest DAP MAE ranging from 34.29 to 38.68 days across M1–M5. In M5, grounding information reduced all models' MAE to near zero, except for the Gemma 4 31B model, which showed 1.47 days. The Qwen3.5 4B and 9B models showed higher MAE values than $B_{med}$, and the Gemma 4 2B model showed DAP MAE values close to $B_{med}$.

Plant count MAE values showed interactions between model size and in-context learning methods. All models with M1–M4 showed higher MAE values than $B_{med}$. Also, only Gemma 4 31B and Qwen3.5 27B models showed lower MAE than $B_{rnd}$ for M1 and M2. Larger Gemma 4 models showed lower plant count MAE values than smaller models, but adding few-shot examples increased the MAE values for M3 and M4 unexpectedly. Models achieved a plant-count error of less than 0.97 plants using the grounding hint in M5.

The plant location CD showed that the Gemma 4 2B and 4B models performed no better than $B_{rnd}$ for all in-context learning methods. The Qwen3.5 4B model and 9B model showed higher error than $B_{rnd}$ for M1 and M2, and the Qwen3.5 9B model showed lower error than $B_{rnd}$ for M3, M4, and M5. The Gemma 4 31B model showed the lowest error for M1, M2, M3, and M4, except for M5 where the Qwen3.5 27B performed better on plant location estimation.

Notably, adding few-shot JSON or image examples often failed to improve geometric accuracy. Our results demonstrate that additional context could reinforce prompt priors more strongly than it improved visual extraction. Furthermore, even if relative plant locations were given at M5, lightweight models could not eliminate the error completely because of arithmetic errors or failure to extract information from the context.

**Lighting and Biophysical Evaluations:** As shown in Figure 3 (f), the Gemma 4 models had lower sun elevation MAE than the Qwen3.5 models from M1 to M4. The Gemma 4 4B and 31B models showed lower MAE values than $B_{med}$ across all methods. In contrast, small models in Qwen3.5, such as 4B and 9B models, showed higher MAE values than $B_{med}$ or $B_{rnd}$, and only the 27B model showed an MAE lower than $B_{med}$ in M4 and M5. Providing sun elevation in the context eliminated the errors, except for the Gemma 4 2B model with M5. The sun azimuth MAE results showed that the Gemma 4 and Qwen3.5 models had similar MAE from M1 to M4. Sun azimuth MAE decreased as Gemma 4 model size increased, and the largest Qwen3.5 model showed lower error than smaller models. Similar to sun elevation, adding a grounding hint to the prompt in M5 decreased the MAE below 17.96°. However, adding few-shot JSON examples or JSON-image pairs did not substantially decrease sun elevation and azimuth MAE.

Chlorophyll MAE results showed that small models in Gemma 4 and Qwen3.5 could not estimate values below $B_{med}$. Gemma 4 models showed decreasing MAE when model size increased, while the Qwen3.5 9B model showed higher MAE than the 4B model in M3 and M4. Adding few-shot JSON and image decreased chlorophyll MAE for the Qwen3.5 27B model in M3 and M4, and the Gemma 4 31B model showed similar MAE for all in-context learning methods. Since chlorophyll content was not included in M5, the grounding hint did not substantially decrease MAE except for the Gemma 4 2B model.

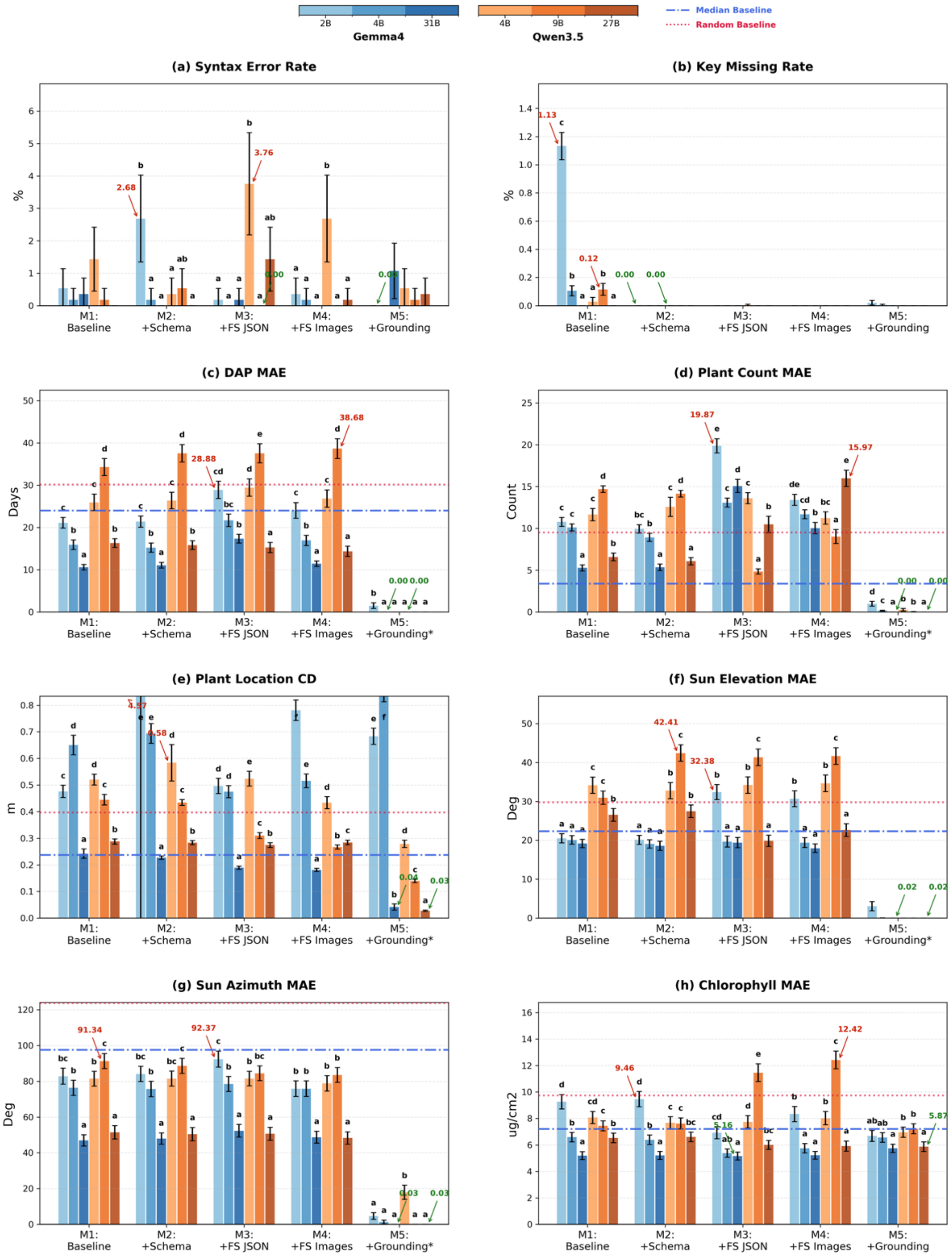

Figure 3. Performance comparison of Gemma 4 (blue) and Qwen3.5 (orange) model families on a synthetic dataset. Panels illustrate (a) syntax error rate, (b) key-missing rate, (c) DAP MAE, (d) plant count MAE, (e) plant location CD, (f) sun elevation MAE, (g) sun azimuth MAE, and (h) chlorophyll MAE. M1–M5 denote progressive in-context learning conditions. For the M5 condition, an asterisk (*) indicates +Grounding, where ground-truth values were provided as inputs for the specified parameters. Error bars represent 95% confidence intervals (CI). Letters indicate significant differences ($p < 0.05$) based on a compact letter display. Green and red values highlight the lowest and highest results within each panel, respectively. Unlabeled bars or empty spaces denote an error value of exactly 0.

**Effect of Dataset DAP:** Figure 4 and Figure 5 show the synthetic data evaluation result of the largest model per model family, grouped by dataset DAP. Based on Figure 4(a) and Figure 5(a), the Qwen3.5 27B model showed a maximum JSON syntax error rate of 3.6%, while Gemma 4 31B model showed maximum 1.8%. Both models showed 0.0% key-missing errors across all methods.

As shown in Figure 4(c) and Figure 5(c), DAP MAE showed a pattern increasing from DAP 10 to 30, decreasing from 30 to 70, and increasing again from 70 to 90 for both models. For 30 DAP with M1, M2, and M3, the Qwen3.5 model showed errors close to $B_{rnd}$, and the Gemma 4 model showed errors lower than $B_{med}$, except for 30 DAP with M3. The MAE for plant count increased for both models as DAP increased, especially for M3 and M4. The Qwen3.5 model showed errors higher than $B_{rnd}$ when DAP was larger than 30 with M3 and M4.

Plant location CD increased as DAP progressed, and adding few-shot JSON and images decreased the plant location error of the Gemma 4 31B model across all DAP, while adding few-shot examples decreased plant location error at 10 and 30 DAP for the Qwen3.5 model. Sun elevation and chlorophyll MAE values increased as the DAP progressed, but sun azimuth MAE values remained similar across DAP.

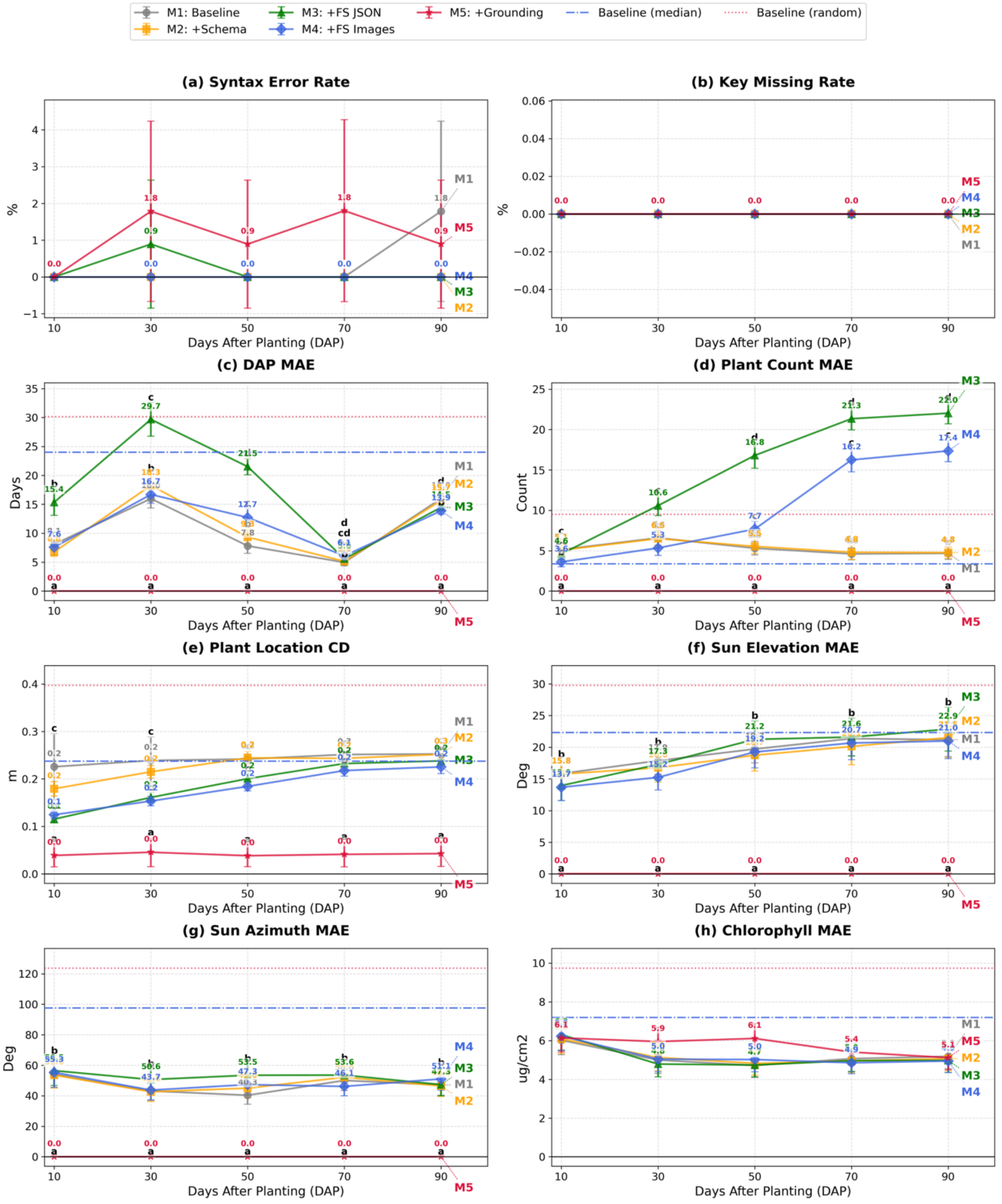

Figure 4. Temporal dynamics of model performance across growth stages (10–90 DAP) of Gemma 4 31B on a synthetic dataset. Panels illustrate (a) syntax error rate, (b) key-missing rate, (c) DAP MAE, (d) plant count MAE, (e) plant location CD, (f) sun elevation MAE, (g) sun azimuth MAE, and (h) chlorophyll MAE. M1–M5 denote progressive in-context learning conditions, represented by distinct colored lines. For the M5 condition, an asterisk (*) indicates +Grounding, where ground-truth values were provided as inputs for the specified parameters. Baselines (median and random heuristics) are indicated by horizontal dash-dotted and dotted lines, respectively. Error bars represent 95% confidence intervals (CI). Black letters above the data points indicate significant differences ($p < 0.05$) among methods at each specific DAP based on a compact letter display.

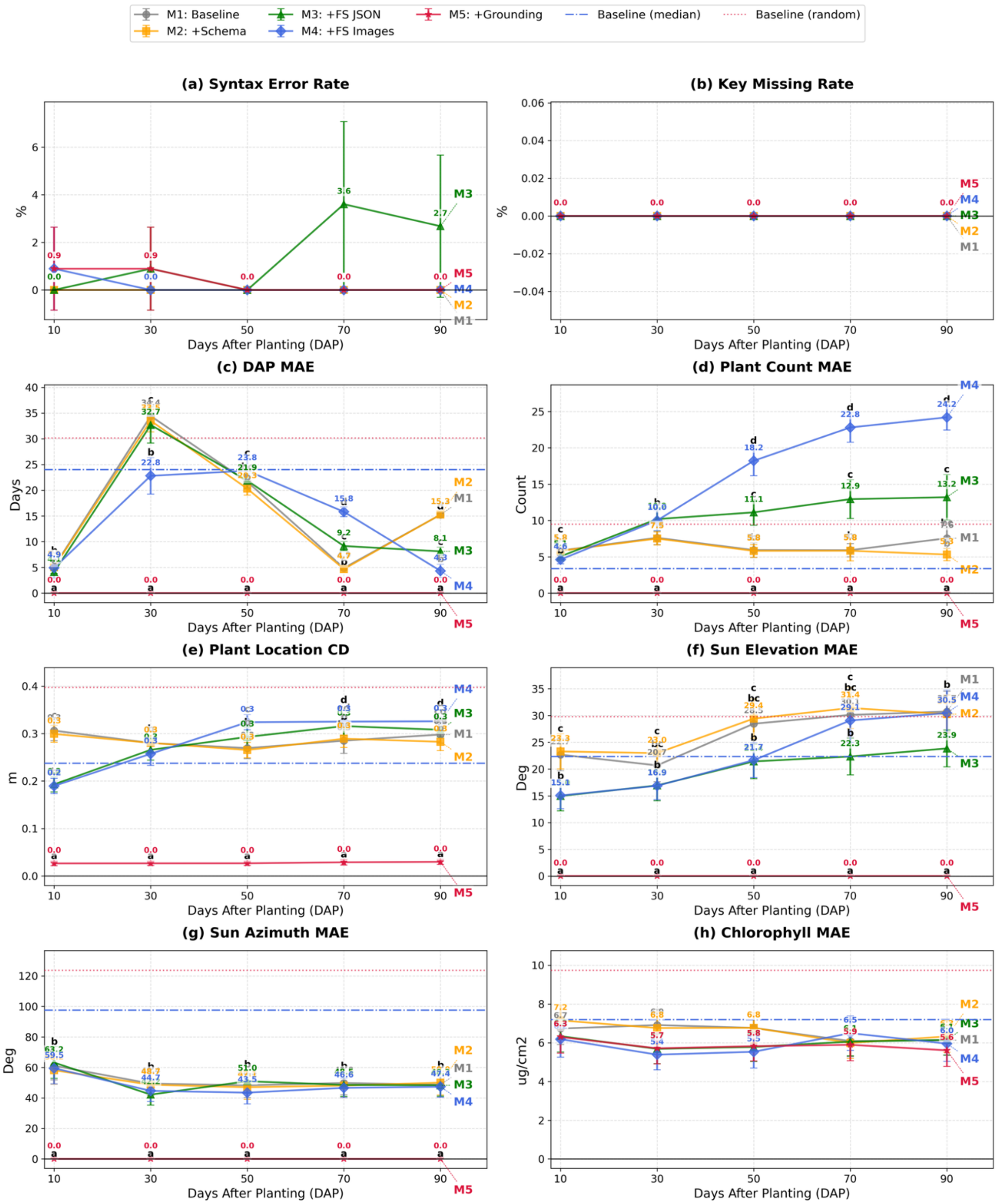

Figure 5. Temporal dynamics of model performance across growth stages (10–90 DAP) of Qwen3.5 27B on a synthetic dataset. Panels illustrate (a) syntax error rate, (b) key-missing rate, (c) DAP MAE, (d) plant count MAE, (e) plant location CD, (f) sun elevation MAE, (g) sun azimuth MAE, and (h) chlorophyll MAE. M1–M5 denote progressive in-context learning conditions, represented by distinct colored lines. For the M5 condition, an asterisk (*) indicates +Grounding, where ground-truth values were provided as inputs for the specified parameters. Baselines (median and random heuristics) are indicated by horizontal dash-dotted and dotted lines, respectively. Error bars represent 95% confidence intervals (CI). Black letters above the data points indicate significant differences ($p < 0.05$) among methods at each specific DAP based on a compact letter display.

## 4.2. Fine-tuned model evaluation for synthetic and real datasets

The effect of synthetic dataset supervision was tested by fine-tuning the largest Gemma 4 and Qwen3.5 models on synthetic data and evaluating them on synthetic and real datasets. Table 2 shows the evaluation metrics for DAP, plant count, plant location, sun elevation, sun azimuth, and chlorophyll of the fine-tuned models. During evaluations, few-shot examples were sampled from the same domain as the evaluation set.

The results show that fine-tuning on the synthetic dataset generally increased correlation coefficients and reduced MAE and sMAPE for the Gemma 4 31B model. For example, fine-tuning on the synthetic dataset increased the correlation coefficient for plant count, solar positions, and chlorophyll, while decreasing MAE and sMAPE for DAP and chlorophyll. Although the Gemma 4 31B improved modestly on the synthetic evaluation, these gains did not transfer to the real dataset, showing lower correlation coefficients and higher MAE and sMAPE values than the original model on real domain evaluation.

Fine-tuning the Qwen3.5 27B model showed degraded performance on both synthetic and real evaluations, indicating that supervised fine-tuning on the synthetic dataset may overfit the model to the training dataset and interfere with the model's pre-existing multimodal reasoning capabilities.

Table 2. Effect of synthetic fine-tuning on synthetic and real dataset evaluations. Fine-tuned values are shown with original model scores in parentheses.

| | | ***Gemma 4 31B*** | | ***Qwen3.5 27B*** | |
|---|---|---|---|---|---|
| ***Parameter*** | **Metric** | **SynFT→SynDS** | **SynFT→RealDS** | **SynFT→SynDS** | **SynFT→RealDS** |
| ***DAP*** | r | 0.90 (0.90) | 0.90 (0.93) | 0.47 (0.82) | 0.54 (0.91) |
| | MAE | 10.4 (11.4) | 10.8 (8.7) | 26.7 (14.4) | 22.2 (9.3) |
| | sMAPE | 18.9% (27.5%) | 21.8% (23.3%) | 45.4% (28.3%) | 42.0% (20.8%) |
| | r | 0.43 (0.32) | 0.17 (0.20) | 0.25 (0.28) | 0.13 (0.20) |
| ***Plant Count*** | MAE | 9.1 (10.0) | 2.8 (2.7) | 9.7 (16.0) | 2.9 (3.0) |
| | sMAPE | 43.3% (50.8%) | 20.8% (20.0%) | 48.6% (79.4%) | 21.3% (22.8%) |
| ***Sun Elevation*** | r | 0.64 (0.54) | 0.33 (0.49) | 0.38 (0.55) | 0.04 (0.20) |
| | MAE | 21.1 (18.0) | 8.8 (7.4) | 27.9 (22.6) | 14.2 (10.6) |
| | sMAPE | 90.4% (53.0%) | 24.9% (16.0%) | 121.3% (86.3%) | 36.1% (21.7%) |
| ***Sun Azimuth*** | r | 0.60 (0.56) | 0.24 (0.53) | 0.39 (0.57) | 0.01 (-0.15) |
| | MAE | 63.2 (64.5) | 14.6 (8.6) | 82.8 (66.4) | 12.7 (14.5) |
| | sMAPE | 59.9% (47.9%) | 13.2% (7.9%) | 51.4% (52.8%) | 11.1% (13.8%) |
| ***Chlorophyll*** | r | 0.77 (0.70) | 0.73 (0.81) | 0.61 (0.55) | 0.23 (0.73) |
| | MAE | 4.3 (5.2) | 19.0 (11.8) | 6.5 (5.9) | 37.6 (10.9) |
| | sMAPE | 32.1% (41.0%) | 37.0% (29.7%) | 50.3% (45.0%) | 65.8% (27.0%) |
| ***Improvements / Total*** | | **12/15** | **1/15** | **4/15** | **5/15** |

## 4.3. Effect of Few-Shot Modality on Real Dataset Evaluation

Table 3 compares the correlation coefficient, MAE, and sMAPE for DAP, plant count, sun elevation, sun azimuth, and chlorophyll between zero-shot, synthetic few-shot (Syn FS), and real few-shot (Real FS) settings. As shown in Table 3, providing synthetic few-shot examples to the Gemma 4 31B model improved the average correlation coefficient from 0.48 to 0.53 but increased average sMAPE from 32.4% to 45.4%, compared to the zero-shot setting (M2). Similarly, the Qwen3.5 27B model also increased the correlation coefficient from 0.25 to 0.49 but also increased sMAPE from 40.2% to 46.5%.

Compared to synthetic few-shot examples, adding real-domain few-shot examples improved the average sMAPE from 45.4% to 19.4% for the Gemma 4 31B model and from 46.5% to 21.2% for the Qwen3.5 27B model. For Gemma 4 31B, the correlation coefficient also rose from 0.53 to 0.59 (Qwen3.5 27B: 0.49 to 0.38). The best overall result was from Gemma 4 31B with real few-shot, achieving a correlation coefficient of 0.59 and an sMAPE of 19.4%.

The DINOv3 VFM-based linear model could estimate the real dataset's DAP with a correlation coefficient of 0.72 and an MAE of 17.9 days with synthetic few-shot images. However, the VFM showed low correlation coefficients for plant count (0.04), sun azimuth (0.04), and chlorophyll (-0.39). When using real few-shot images in VFM linear model fittings, the average correlation coefficient increased from 0.15 to 0.46 and decreased the average sMAPE from 55.6% to 26.1% compared to synthetic few-shot fittings.

Table 3. Effect of few-shot data modality for the real dataset evaluation. DINOv3 embedding-based linear model was evaluated as a baseline.

| | | ***Gemma 4 31B*** | | | ***Qwen3.5 27B*** | | | ***DINOv3 SAT*** | |
|---|---|---|---|---|---|---|---|---|---|
| ***Parameter*** | **Metric** | **No FS (M2)** | **Syn FS (M4)** | **Real FS (M4)** | **No FS (M2)** | **Syn FS (M4)** | **Real FS (M4)** | **Syn FS (M4)** | **Real FS (M4)** |
| ***DAP*** | r | 0.81 | 0.88 | 0.93 | 0.88 | 0.86 | 0.91 | 0.72 | 0.75 |
| | MAE | 14.0 | 10.9 | 8.7 | 11.3 | 11.9 | 9.3 | 17.9 | 16.1 |
| | sMAPE | 35.2% | 27.8% | 23.3% | 31.6% | 30.4% | 20.8% | 43.9% | 35.5% |
| ***Plant Count*** | r | 0.20 | 0.10 | 0.20 | 0.09 | 0.14 | 0.20 | 0.04 | 0.01 |
| | MAE | 8.5 | 19.7 | 2.7 | 9.2 | 21.5 | 3.0 | 4.7 | 3.5 |
| | sMAPE | 42.0% | 73.1% | 20.0% | 44.8% | 53.2% | 22.8% | 30.0% | 27.2% |
| ***Sun Elevation*** | r | 0.40 | 0.54 | 0.49 | -0.18 | 0.32 | 0.20 | 0.33 | 0.23 |
| | MAE | 10.9 | 13.0 | 7.4 | 13.3 | 17.4 | 10.6 | 20.6 | 7.6 |
| | sMAPE | 20.9% | 25.4% | 16.0% | 25.0% | 44.1% | 21.7% | 57.2% | 16.5% |
| ***Sun Azimuth*** | r | 0.28 | 0.53 | 0.53 | 0.27 | 0.46 | -0.15 | 0.04 | 0.65 |
| | MAE | 38.6 | 109.7 | 8.6 | 70.0 | 108.2 | 14.5 | 49.7 | 14.6 |
| | sMAPE | 26.3% | 58.9% | 7.9% | 49.1% | 66.8% | 13.8% | 34.8% | 15.0% |
| ***Chlorophyll*** | r | 0.71 | 0.61 | 0.81 | 0.18 | 0.67 | 0.73 | -0.39 | 0.67 |
| | MAE | 15.9 | 18.3 | 11.8 | 24.5 | 15.9 | 10.9 | 34.0 | 13.9 |
| | sMAPE | 37.8% | 41.9% | 29.7% | 50.6% | 37.9% | 27.0% | 112.2% | 36.2% |
| ***Average*** | r | 0.48 | 0.53 | 0.59 | 0.25 | 0.49 | 0.38 | 0.15 | 0.46 |
| | sMAPE | 32.4% | 45.4% | 19.4% | 40.2% | 46.5% | 21.2% | 55.6% | 26.1% |

## 4.4. Visual Quality Evaluation

Figure 6 and Figure 7 show the generated cowpea plot images from the synthetic and real datasets. As shown in Figure 7, the Gemma 4 31B model could reasonably estimate plant locations but had parameter estimation errors. For example, M1–M4 estimated larger plants than the input images, and brighter and yellowish plants because of chlorophyll content and solar position errors. Providing few-shot examples removed overestimated plants on 90 DAP from M1 to M3 but could not lower the average sMAPE substantially.

Similarly, tests on real images generally produced estimates of higher plant DAP, resulting in bigger plants on 10 and 28 DAP (Figure 7). Adding few-shot images (M4) improved the hallucinated plant locations on 28 DAP with M1 to M3, but the effect of few-shot images on later DAP was not visually recognizable. For the real dataset, adding more context from M1 to M5 decreased the sMAPE, showing an average sMAPE of 21.6% for this example.

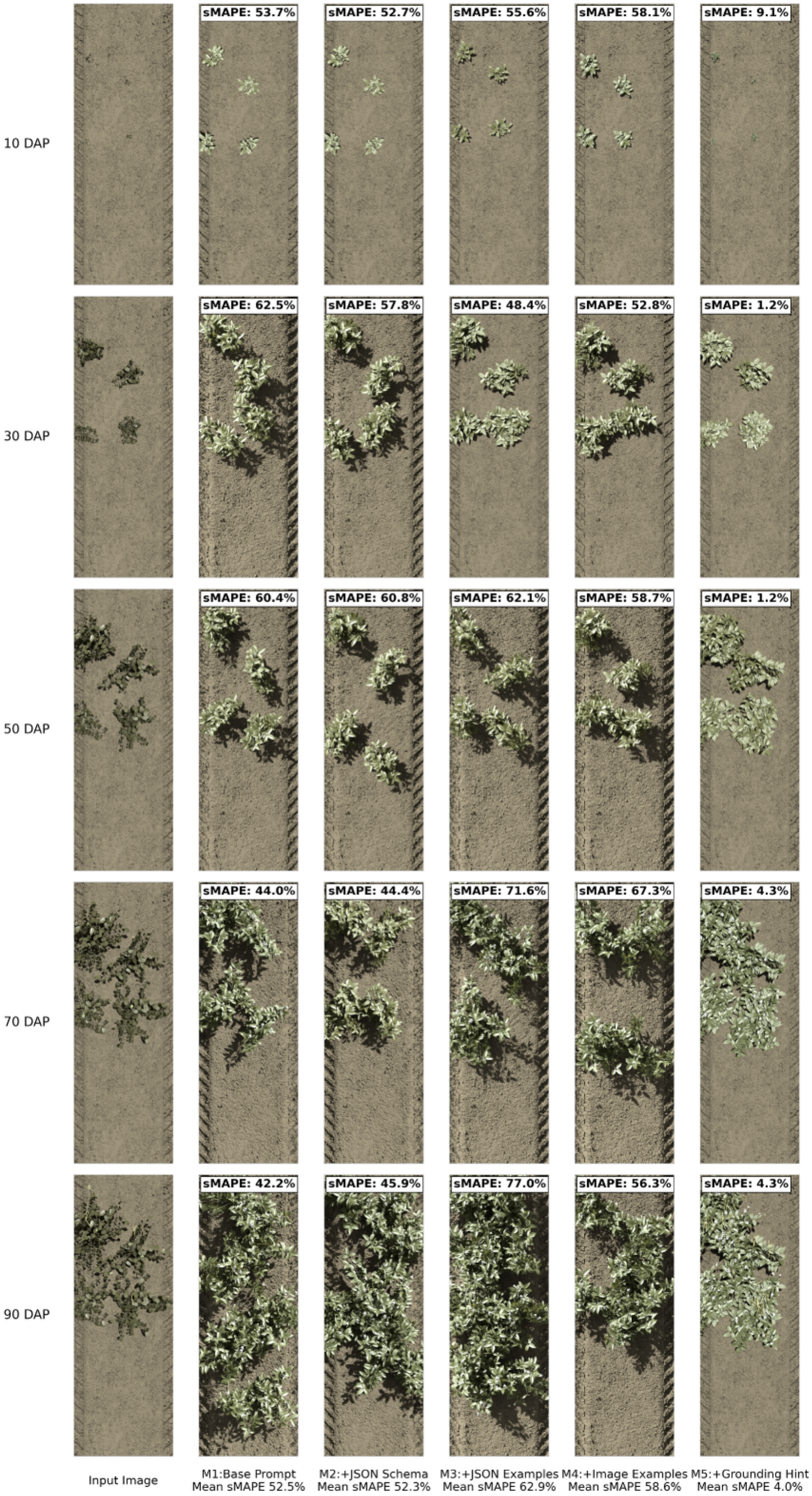

Figure 6. Examples of simulated cowpea plot generation results from the synthetic dataset based on in-context learning methods using the Gemma 4 31B model.

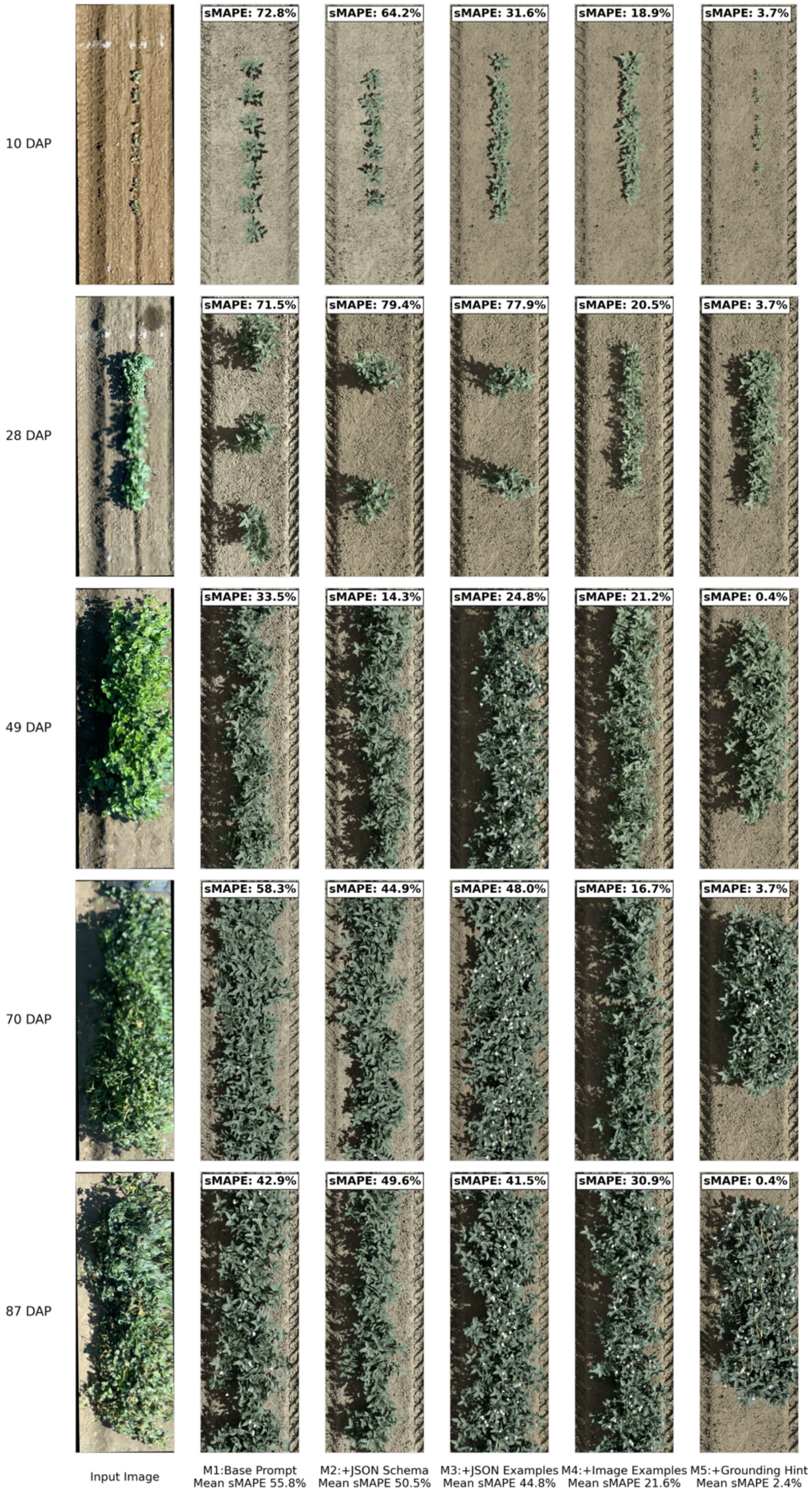

Figure 7. Examples of simulated cowpea plot generation results from the real dataset based on in-context learning methods using the Gemma 4 31B model.

# 5. Discussion

## 5.1. Structured Output Generation Performance

Forcing LLMs to respond in JSON format can be enabled by setting JSON mode in most LLM APIs, which provides a template and generates structured output that follows the given schema. Zhang et al. (2023) argued that structured decoding can improve tool-use performance by removing errors in JSON generation, even without fine-tuning the model. Prior work reported that structured output could not only improve the performance of classification tasks such as information extraction and named entity recognition, but also decrease the number of generated tokens (Beurer-Kellner et al., 2023; Geng et al., 2023). However, forcing an LLM to generate responses in a structured format reduced performance on complex reasoning tasks, such as mathematics, last-letter concatenation, and object shuffling (Tam et al., 2024). Iwanowski and Gahbler (2025) reported that forcing structured output in object localization tasks can lead to hallucinations, generating fake objects that conform to the desired structure.

In Figure 3(a) and (b), our results showed a maximum of 3.76% syntax error and 1.13% JSON key-missing error. Our benchmark shows that syntactic validity and semantic fidelity can diverge; for example, the Gemma 4 31B model showed 1.07% of JSON syntax errors when grounding hints were provided, even though this configuration showed the lowest errors and best visual quality. For practical deployment, structural correctness should therefore be treated as a prerequisite rather than as evidence of the semantic fidelity of the generated result, since perfect structured document generation performance is crucial for integrating the automated plant simulation generation pipeline into LLM agents and tool use (Yang et al., 2026; Zhang et al., 2023).

## 5.2. Effect of Model Size and Contextual Bias

The main finding of this study is that additional context does not reliably improve the in-context learning results. In several configurations, longer prompts increased error, because the models fell back on contextual priors, copying values or distributions from the prompt, rather than extracting evidence from the target image. This effect was especially pronounced in smaller models, but it was not limited to them.

For example, Figure 3(c) shows that increasing model size occasionally increased DAP MAE for certain models, such as Qwen3.5 4B and 9B. This irregularity suggests that model size alone does not guarantee improved accuracy on zero-shot and few-shot agricultural tasks. Furthermore, additional context often increased MAEs, such as Figure 3 (c), (d), (e), and (f), where DAP, plant count, plant location, and sun elevation errors increased after providing few-shot JSON or image examples. This suggests small models, such as Gemma 4 2B or Qwen3.5 9B, are unable to extract information from the target image reliably when the model's context increases, causing models to copy the values from the instruction prompt or few-shot examples, increasing the error metrics. The increase in error suggests that few-shot examples can act as noise when the model fails to capture a reliable visual signal and instead focuses on the wrong contextual patterns.

## 5.3. Effect of Fine-Tuning

Our fine-tuning results indicate that synthetic supervision alone is not sufficient to improve the robustness of simulation parameter estimation. As shown in Table 2, while Gemma 4 31B benefited modestly from fine-tuning on synthetic evaluation, these gains did not transfer to real-domain evaluations, and Qwen3.5 27B degraded after fine-tuning in both domains. This result suggests that fine-tuning can amplify the domain gap or disrupt capabilities that support zero-shot inference of VLMs. Luo et al. (2025) reported that continual fine-tuning can introduce catastrophic forgetting, particularly in larger models. Several studies reported the risk of performance degradation of multimodal large language models during fine-tuning (Li et al., 2024; Tang et al., 2025; Zhu et al., 2024). Further research is needed to find a better fine-tuning method to improve the accuracy of plot parameter estimation for both the synthetic and real domains without catastrophic forgetting.

### 5.4. Effect of Few-Shot Modality

As shown in Table 3, the modalities of the few-shot examples mattered substantially. Real few-shot examples generally outperformed synthetic few-shot examples on the real-dataset evaluation across both VLM families, indicating that prompt-domain alignment is important for maximizing in-context learning performance.

Similarly, the DINOv3 VFM showed better performance when the linear model was fitted from real few-shot examples and evaluated on the real dataset. However, the VFM fitted with synthetic few-shot examples could estimate DAP of real data but failed to estimate chlorophyll. Compared to VFMs' weak few-shot learning ability across domains, providing synthetic few-shot examples to VLMs led to performance improvement in DAP, solar positions, and chlorophyll.

This benefit will be useful when accurate ground truth is difficult to measure or inaccessible but synthetically guided context can be given, such as canopy photosynthesis, growth, water use, biomass estimation, and detailed plant architecture.

### 5.5. Integration with the Digital Twin Pipeline and Practical Applications

Digital twins simulate physical systems and update their states based on sensor data, enabling AI-driven plant simulations and 'what-if' experiments across crops and environments (Ganapathysubramanian et al., 2025; Verdouw et al., 2021). Our method bridges the gap between physiological metrics used by agronomists and morphological observations used by growers (Archambault et al., 2024), by simultaneously generating morphological traits (e.g., plant count, locations) and biophysical parameters, both encoded as JSON files and visualized through 3D plot simulations.

This approach aligns with the multi-agent digital twin architecture proposed by Skobelev et al. (2020), where agents interact with agronomists through mobile interfaces to continuously recalculate yield forecasts and generate recommendations. Although our current results showed 19.4% average sMAPE for plant parameter estimation with few-shot examples (Table 3, Gemma 4 31B with Real FS (M4)), further improvements can be achieved by optimizing few-shot examples with accurately measured parameters and images. By combining in-context learning with precisely calibrated few-shot examples, we minimize sampling requirements to the few-shot scale (i.e., up to 5 samples) rather than the fine-tuning dataset scale (i.e., more than 1,000 samples).

## 6. Conclusion

We have introduced a benchmark for evaluating whether VLMs can generate structured plant simulation configurations from images and studied the effects of in-context learning methods, model family, model size, fine-tuning, and the modality of few-shot examples. The results show that current VLMs can produce syntactically valid JSON outputs and coarse parameter estimates, but performance remains inconsistent across model families and frequently below simple baselines for key geometric and biophysical parameters.

Domain-aligned few-shot examples improved results across both VLM families, yet models remained sensitive to contextual bias, often falling back on prompt priors rather than extracting visual evidence. Furthermore, synthetic fine-tuning improved synthetic-domain performance but failed to transfer to real-domain evaluations, and in some cases degraded performance, suggesting that naïve synthetic adaptation can amplify domain mismatch or disrupt zero-shot inference capabilities.

Therefore, we view image-to-simulation JSON generation as a promising yet challenging task for agricultural VLM research. Future work should explore stronger visual grounding techniques, domain-invariant fine-tuning strategies, and structured decoding methods to improve JSON generation reliability. Expanding the benchmark to additional crop species, incorporating multispectral images, and evaluating under real-world deployment will further clarify the practical pathways for integrating VLMs into automated plant simulation pipelines.

## Acknowledgements

This work was supported, in whole or in part, by the Bill & Melinda Gates Foundation INV-0028630. Under the grant conditions of the Foundation, a Creative Commons Attribution 4.0 Generic License has already been assigned to the Author Accepted Manuscript version that might arise from this submission.